\documentclass[letterpaper, 10 pt, conference]{ieeeconf}  
\usepackage{amsmath,amsfonts}
\usepackage{algorithmic}
\usepackage{algorithm}
\usepackage{array}
\usepackage[caption=false,font=normalsize,labelfont=sf,textfont=sf]{subfig}
\usepackage{textcomp}
\usepackage{stfloats}
\usepackage{url}
\usepackage{verbatim}
\usepackage{graphicx}
\usepackage{cite}
\usepackage{orcidlink}
\usepackage{pifont}
\hypersetup{
colorlinks=true,
linkcolor=black
}
\IEEEoverridecommandlockouts                              

\title{\LARGE \bf
Design of an Adaptive Modular Anthropomorphic Dexterous Hand for Human-like Manipulation
}

\author{Zelong Zhou$^{1~\orcidlink{0009-0006-1755-1409}}$, Wenrui Chen$^{1~\orcidlink{0000-0002-6366-7721
}}$, Zeyun Hu$^{1~\orcidlink{0009-0006-7211-0972}}$, Qiang Diao$^{1~\orcidlink{0000-0002-6545-4340
}}$, Qixin Gao$^{1}$, Cuo Yan$^{1}$, Yaonan Wang$^{1~\orcidlink{0000-0002-0519-6458
}}$
\thanks{
This work is supported by the National Key RD Program under Grant 2022YFB4701400/2022YFB4701404, the National Natural Science Foundation of China under Grant 62273137, the Hunan Science Fund for Distinguished Young Scholars under Grant 2024JJ2027.
}
\thanks{$^{1}$School of Artificial Intelligence and Robotics, Hunan University, Changsha
410012, China (e-mail: rosesaika@163.com; chenwenrui@hnu.edu.cn)
}
}   
\begin{document}

\maketitle

\thispagestyle{empty}
\pagestyle{empty}

\begin{abstract}

Biological synergies have emerged as a widely adopted paradigm for dexterous hand design, enabling human-like manipulation with a small number of actuators. Nonetheless, excessive coupling tends to diminish the dexterity of hands. This paper tackles the trade-off between actuation complexity and dexterity by proposing an anthropomorphic finger topology with 4 DoFs driven by 2 actuators, and by developing an adaptive, modular dexterous hand based on this finger topology. We explore the biological basis of hand synergies and human gesture analysis, translating joint-level coordination and structural attributes into a modular finger architecture. Leveraging these biomimetic mappings, we design a five-finger modular hand and establish its kinematic model to analyze adaptive grasping and in-hand manipulation. Finally, we construct a physical prototype and conduct preliminary experiments, which validate the effectiveness of the proposed design and analysis.

\end{abstract}

\section{INTRODUCTION}
The human hand exemplifies unparalleled dexterity. Comprising 54 of the body’s 206 bones—approximately one-quarter of the skeletal system—and supported by a highly intricate muscular architecture, it affords more than 20 degrees of freedom (DoFs) in the fingers alone \cite{han2024journey}. Replicating such mechanical and functional sophistication, Developing a compact and integrated dexterous hand that faithfully mirrors human morphology while retaining flexibility—without reliance on external actuators—continues to pose a formidable engineering challenge.

Ingram et al.~\cite{ingram2008statistics} observed that the human hand exhibits characteristic motion patterns that can be described as a reduction in effective degrees of freedom during specific tasks. This finding implies that dexterous control may be realized with fewer actuators, without substantially compromising functional capability. Motivated by this principle, recent research has increasingly focused on underactuated dexterous hands. These designs typically maintain the full set of 20 degrees of freedom (DoFs), with the majority of joints controlled through synergistic mechanisms. Such as the Schunk Hand\cite{CERULO201775}, Hannes’s Hand\cite{laffranchi2020hannes}, and ILDA KIM-Hand\cite{kim2021integrated} adopt this architecture, using kinematic synergies to balance dexterity and actuation simplicity. This approach offers an effective trade-off between mechanical efficiency and functional capability, making it suitable for both research and real-world applications.

Among these, self-adaptive mechanisms have emerged as another promising underactuated strategy\cite{negrello2020hands}. By fully leveraging synergies, multiple fingers can conform to objects using only a small number of actuators\cite{liu2016biomechanical}. A notable example is the Hand designed by Ritsumeikan University, which actuates 15 joints using only two motors via mechanically coupled tendons\cite{6697149}. Similarly, Sun\cite{sun2021design} proposed a five-finger design with two actuators, capable of reproducing the two primary human grasp types. Other representative designs such as the , HERI Hand\cite{ren2017heri}, Columbia Hand\cite{wang2011highly} and Chen Hand\cite{chen2021design}—adopt adaptive underactuated structures to achieve stable grasping with simplified actuation. However, reliance on complex coupling mechanisms often compromises finger-level manipulability and mechanical performance. The absence of independently controllable degrees-of-freedom significantly limits their effectiveness in tasks requiring manipulation or operation in unstructured environments\cite{odhner2015stable}.

In this study, we explores the feasibility of a balanced design for dexterous robotic hands by leveraging the principles of underactuation and structural adaptability. Focus on replicating the dexterity of the human hand with fewer actuators. Informed by the analysis of the structural and kinematic characteristics of the human hand, we propose a modular dexterous hand structure that retains intra-finger synergies to balance dexterity and actuator efficiency. A modular finger design is presented which achieves four degrees of freedom while requiring only two actuators. In this design, the flexion–extension (FE) and adduction–abduction (AA) motions are consolidated within a single MCP structure, with each motion mode powered by dual-actuator coordination. By embedding a compliant underactuated transmission mechanism, the three-joint finger achieves variable stiffness modulation and adaptive grasping. Finally, a dexterous hand prototype is constructed based on the proposed model. Preliminary dexterous manipulation experiments were conducted to validate the proposed design and confirm its capability to achieve human-like grasping and in-hand manipulation.

\section{Biomimetic Design of the Modular Hand}

\subsection{Skeletal Structure}

\begin{figure}[h]
    \centering
    \includegraphics[width=1\linewidth]{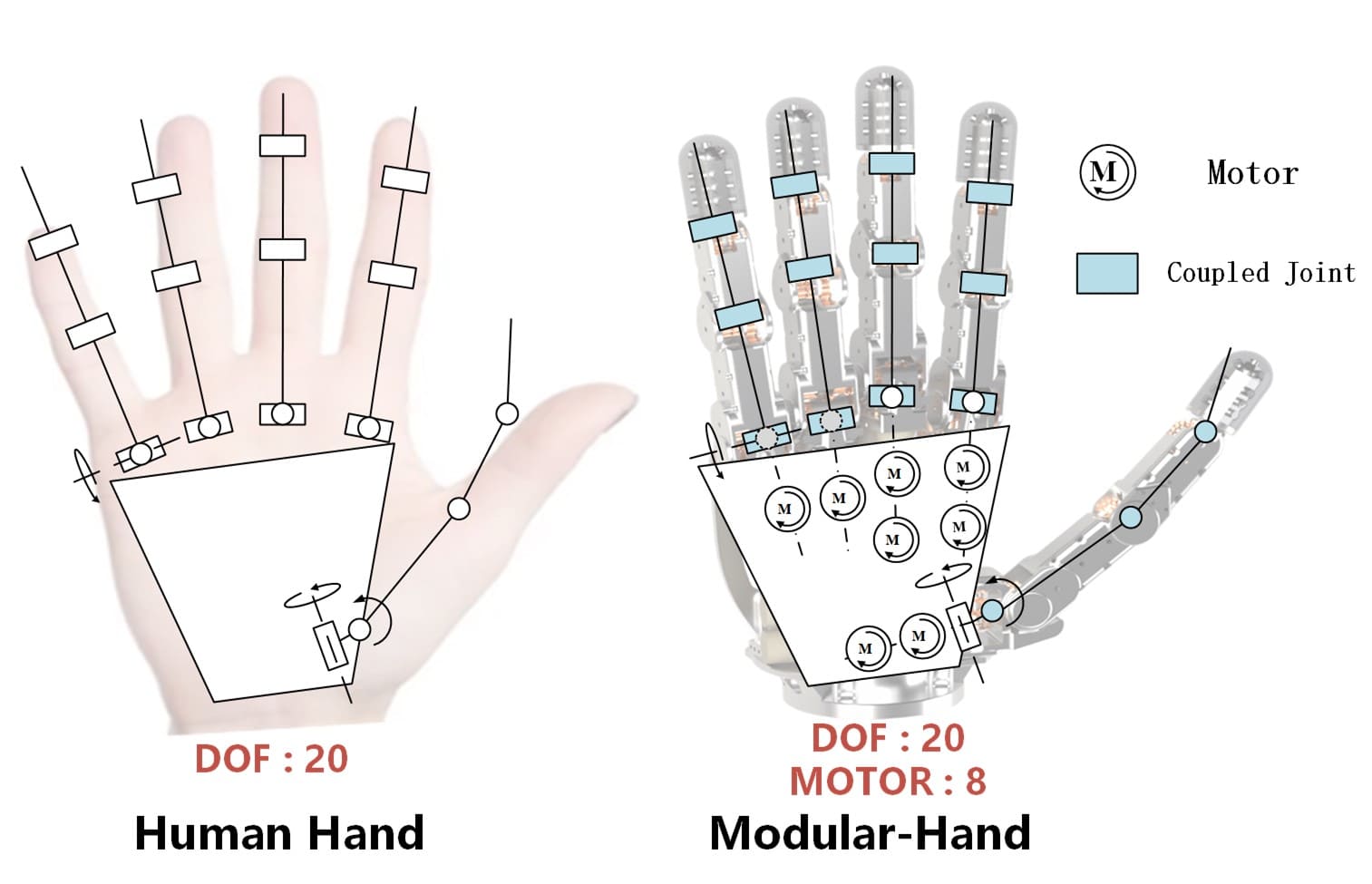}
    \caption{Comparison between the human hand configuration and the modular five-finger hand configuration}
    \label{fig:configuration}
\end{figure}

The human hand is the foundation for anthropomorphic dexterous hand design \cite{piazza2019century}, and has been extensively analyzed from multiple perspectives. Biomechanics studies indicate that the skeletal structure provides the physical support of the human hand\cite{spalteholz2013atlas}. Ligaments and muscles interconnect the structure, forming 15 articulating finger joints with over 20 degrees of freedom. Their motions can be represented as axial rotations about joints\cite{bullock2012assessing}, as shown in Fig.~\ref{fig:configuration}. From a kinematic perspective, the human hand is characterized by its motions and gestures. Cutkosky’s taxonomy\cite{cutkosky1989grasp} provides a qualitative classification of human grasp postures, establishing a macro-level dichotomy of grasp types. Feix’s taxonomy\cite{feix2015grasp} refines the granularity to 33 grasp types, yielding a comprehensive database of human grasp behaviors closely aligned with real-world tasks. Complementing these qualitative frameworks, Santello et al.\cite{santello1998postural} performed a statistical analysis of joint-angle relationships, quantitatively capturing the coordination patterns across the hand. Collectively, these studies reveal strong regularities in human hand structure, offering critical insights for guiding dexterous hand design. These regularities are reflected in the following aspects:

\textit{Coupled joint motion during flexion–extension.} Santello showed via principal component analysis that joint angles along a finger maintain proportional relationships during flexion, making the overall finger posture more critical than individual joint motion. Kamper~\cite{kamper2003stereotypical} showed that fingertip trajectories exhibit stable and predictable spatial patterns, which can be approximated by a single-variable logarithmic spiral, thus capturing the motion as a single degree-of-freedom representation.

\textit{Auxiliary role of the ring and little fingers.} In Feix’s 33 grasp types , the thumb is used in 32, the index finger in all 33, and the middle finger in 28. Thumb–index–middle combinations occur in 27 types, with at least two of these three involved in every grasp. For small-object manipulation, only the thumb, index, and occasionally the ring finger are typically engaged, while the structurally weaker little finger plays a minor role.\cite{zheng2022human}

\textit{Importance of lateral motion in primary digits.} Abduction–adduction of primary fingers enables adjustable inter-finger spacing, expanding fingertip workspace and supporting stable grasps on asymmetrical or oblique objects\cite{yan2021influence}. Okamura et al.\cite{okamura2000overview} identified redundant spatial degrees of freedom as essential for dexterous manipulation, with independent lateral motion being a key capability retained for functional digits.

\begin{figure}[h]
    \centering
    \includegraphics[width=1\linewidth]{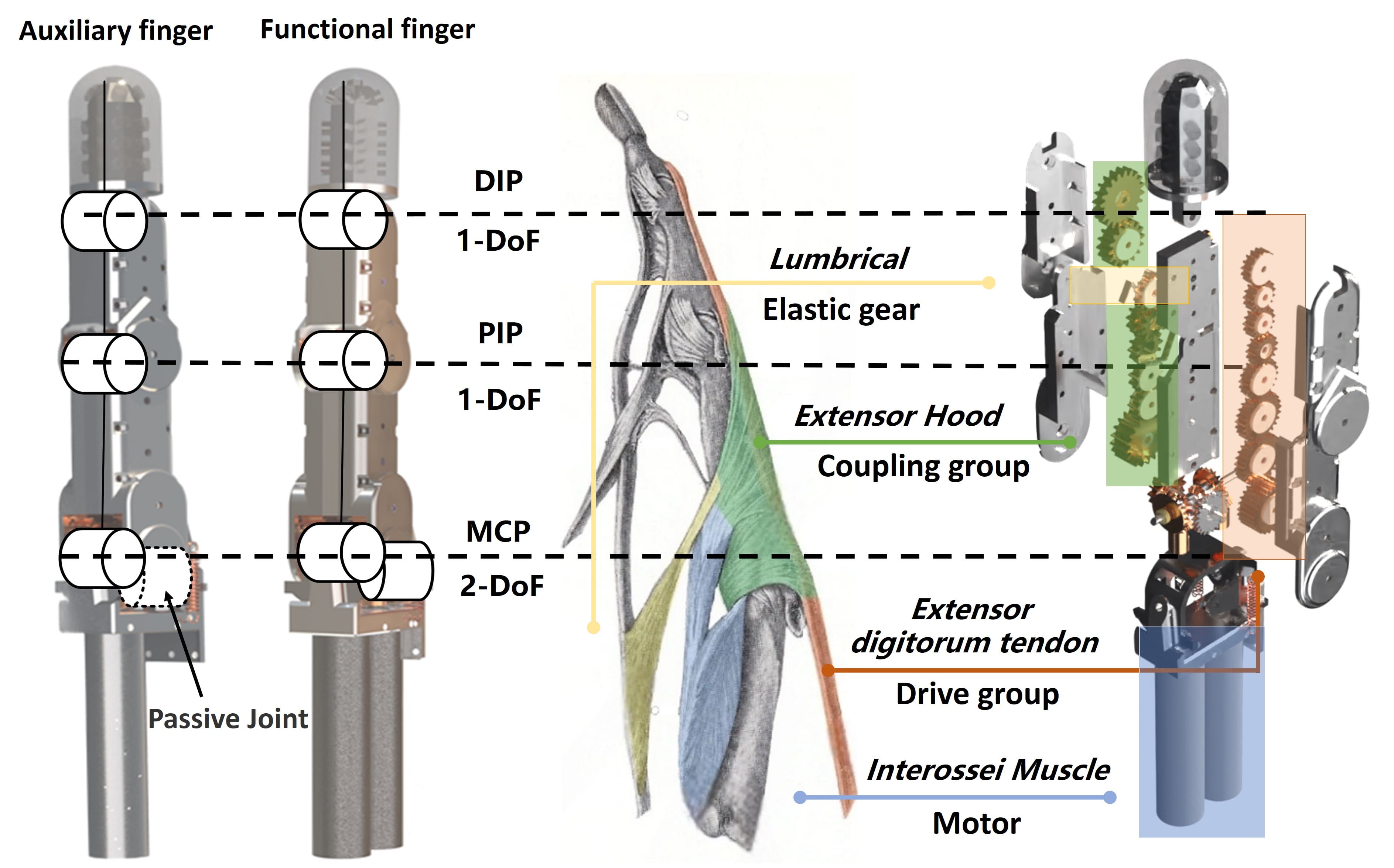}
    \caption{Illustration of the biomimetic mechanical configuration of the modular finger}
    \label{fig:placeholder}
\end{figure}

Based on the above observations, we propose a modular finger configuration, as shown in Fig.~\ref{fig:placeholder}. The modular finger closely resembles the human finger, featuring three joints and four degrees of freedom. Motor–gear assemblies are employed to emulate human tendon actuation, reproducing both flexion–extension and abduction–adduction motions while preserving the natural joint coupling. Following the principle of replicating human hand performance with minimal actuation, the ring and little fingers are simplified as auxiliary fingers, retaining only passive lateral motion capability. The manipulator configuration based on modular fingers is shown in Fig.~\ref{fig:configuration}.

\subsection{Driven System}

In dexterous hand design, tendon-driven mechanisms are widely employed as biomimetic elements due to their compactness, light weight, and passive flexibility\cite{zhang2025dexterous}. However, inherent elasticity degrades both static and dynamic performance, while pretension loss from material fatigue, environmental influences, and mechanical wear undermines precision and repeatability\cite{agrawal2010modeling}\cite{gao2017general}. To meet the growing demands of complex applications, a mechanically optimized solution is required to balance dexterity, stability, and cost efficiency. We introduce a hybrid gear–elastic transmission that emulates human muscle actuation and protective adaptability while shortening the transmission chain to the MCP-DIP segment. This design enhances stability and control accuracy, effectively addressing the limitations of conventional tendon-based approaches.

\begin{figure}[h]
    \centering
    \includegraphics[width=1\linewidth]{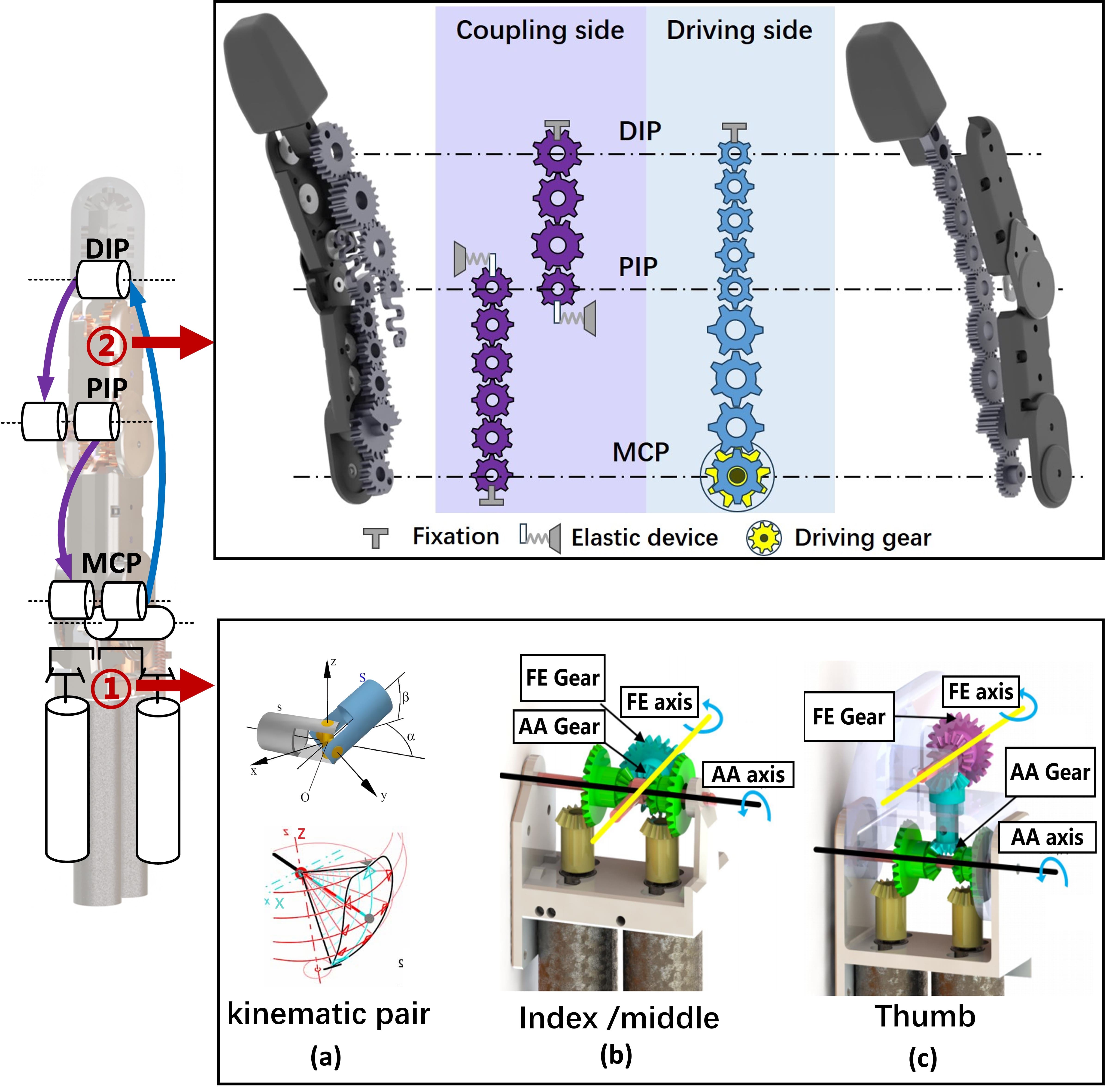}
    \caption{(1). Modular finger transmission chain with adaptive coupling capability; (2). Comparison between the human hand configuration and the modular five-finger hand configuration}
    \label{fig:mcp}
\end{figure}

According to the study of human skeletal joints by Nierop \cite{van2008natural}, the equivalent kinematic pair of the MCP joint is illustrated in Fig.~\ref{fig:mcp}(a). The proximal joint functions as a compound joint, integrating both flexion–extension and lateral swing, thereby requiring actuation to support both motions. In this work, a gear differential device is employed at the proximal joint as the driving mechanism. By leveraging dual motor inputs, the device enables coordinated joint control through differential transmission, achieving finger flexion–extension and lateral swing as a linear combination of motions, as depicted in Fig.~\ref{fig:mcp}(b)(c). For auxiliary fingers, the differential coupling mechanism is omitted; however, lateral swing capability is preserved through a spring-based passive mechanism, which provides compliant side motion and facilitates auxiliary object envelopment during grasping tasks.

 The knuckle transmission system employs a dual-gear configuration with a driving side and a coupling side, as shown in Fig.~\ref{fig:mcp}. The driving gear chain delivers power, while the coupling gear group enforces inter-joint motion coupling. Based on the collaborative bionic motion analysis in the preceding section, maintaining humanoid dexterity requires strong inter-joint coupling and correlated motion. This is achieved by a two-layer gear chain that couples the three flexion–extension joints, enabling coordinated three-DOF motion.


Conventional rigidly coupled gear sets can synchronize three joints but lack passive adaptability to object geometry. To address this issue, power is transmitted from the distal to the proximal knuckle through the driving-side gears, while elastic gears are introduced in the coupling group to replace selected fixed gears. For example, the proximal–middle knuckle pair is linked by an elastic element. Without external forces, the elastic gear acts as a fixed gear, preserving rigid coupling. When the proximal joint encounters resistance, the elastic element deforms, breaking the rigid coupling. Power continues to the middle phalanx, which then overcomes the elastic resistance to adaptively conform to the object.

\section{Dexterity and Compliance Analysis}
\subsection{Workspace Analysis}
\begin{figure*}[b]
    \centering
    \includegraphics[width=1\linewidth]{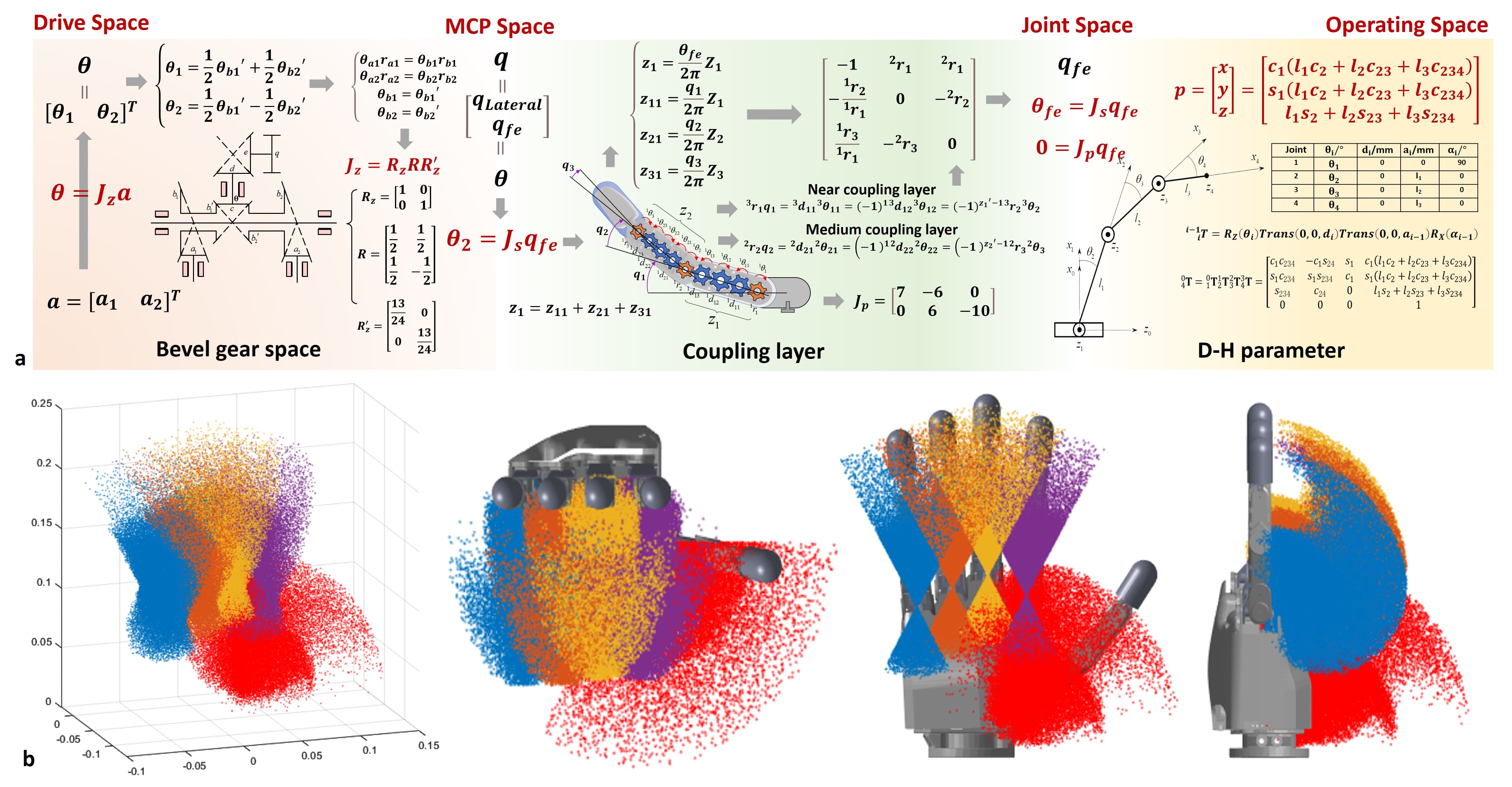}
    \caption{(a) Schematic diagram of the computational process from the driving space to the workspace; (b) Visualization of the workspace using the Monte Carlo method and its projections on the XOY, YOZ, and XOZ planes.}
    \label{fig:workplace}
\end{figure*}

The motion of humanoid dexterous hand adopts modular design, which is composed of five modular fingers. The overall mechanical structure, motion relationship and working space are determined by each modular finger, so the kinematics analysis of the finger is very important. We analyze the driving space, joint space and operation space of the finger and the kinematics between them to verify the advantages of the proposed structure.The calculation process is shown in Fig~\ref{fig:workplace}.

The modular finger is driven and controlled by two motors, and the driving space is expressed as:
$$a=[a_1,a_2 ]^T$$

The modular finger has four joints, of which the proximal joint is a composite joint of flexion and extension and side swing. The finger has four degrees of freedom of motion, and the joint space can be expressed as joint angle :
$$q=[q_{aa},q_1,q_2 ,q_3 ]^T$$

According to the kinematics of the planar mechanism, the motion relationship between the gears of the differential coupling device is calculated. The planetary gear has rotation and revolution motion, and its motion characteristics are expressed as :
$$\theta=[\theta_1,\theta_2]^T$$

According to the relationship between gear radius and tooth number, the kinematic relationship of gear differential mechanism can be described as :
$$\theta=J_z a$$

Among them, $R_z$ is the transmission coefficient from gear $c$ to the output end, $R$ is the differential coupling relationship from the inner side of two double-layer bevel gears to gear $c$, and $R_z^\prime$ is the transmission coefficient from two motor gears to the inner side of two double-layer bevel gears. 

The transmission coefficients are :
$$R_z=\begin{bmatrix}
  1&0 \\
  0&1
\end{bmatrix},
R=\begin{bmatrix}
  \frac{1}{2} &\frac{1}{2} \\
  \frac{1}{2}&-\frac{1}{2}
\end{bmatrix},
R_z\prime=\begin{bmatrix}
  \frac{13}{24} &0\\
  0&\frac{13}{24}
\end{bmatrix}$$

The side swing action of the modular finger is mapped to the revolution of the planetary gear, so :
$$q=\theta$$

where:
$$q=\begin{bmatrix}
  q_{aa}\\
  q_{fe}
\end{bmatrix}$$

So far, the mapping from the driving space to the near joint space is realized. 

The driving group gears are actually rotating and rotating around the axis at the same time. For the three-joint driving gear, the rotation angle can be decomposed into the joint flexion and extension angle and the power transmission angle. After plane mechanics calculation, the relationship between the rotation angle of the near-joint drive space and the joint flexion and extension space can be obtained as follows : 
$$\theta_{2}=J_s q_{fe}$$
where:$$q_{fe}=\begin{bmatrix}q_1&q_2&q_3\end{bmatrix}^{T}  $$

The number of coupling gear teeth at each joint is shown in Table~\ref{tab:table1}. 

\begin{table}[h]
\centering
\caption{Teeth Number of three joint drive gears\label{tab:table1}}
\begin{tabular}{|c|c|c|c|}
\hline Joint & MCP&PIP & DIP  \\ \hline
Teeth   & 22($z_1$)  & 20($z_2$)  & 16$(z_3)$   \\ \hline
\end{tabular}
\end{table}
The ratio of near, middle and far teeth is 14 : 12 : 20, and the coupling ratio is 6 : 7 : 4.2.  The coupling transmission coefficient ${J_s}$  can be obtained as :
$$J_s=\begin{bmatrix}7&-6&0\\ 0&6&-10 \end{bmatrix}$$

Based on the mapping from the driving space to the joint space, the kinematics analysis of the finger is carried out by the D-H parameter method. The finger shell is composed of rigid parts. From the overall point of view of the finger movement, each knuckle is similar to the connecting rod, and the modular finger is similar to the articulated serial robot. The Monte Carlo method is used to express the workspace of the modular finger\cite{chaudhury2018workspace}. The visual expression of the workspace of the dexterous hand is shown in Fig~\ref{fig:workplace} (b). We map the workspace to the dexterous hand simulation model. According to the projection of the finger workspace in the XOY, XOZ and YOZ planes, it is found that the modular finger has a rich in-hand operating space in the five-finger hand model. 

\subsection{Analysis of Underactuated Compliant Mechanism(UCM)}
In the transmission structure of the finger, the mechanical flexibility is arranged in series with the actuator, which improves the stability and impact resistance during the grasping and dexterous tasks. However, when there exist uncontrollable motion and forces\cite{prattichizzo2013motion}, which may lead to undesirable results, affecting the grasping and operation of the dexterous hand, such as the s ejection phenomenon \cite{birglen2003force}, that is, the grasp sequence can degenerate into ejecting the object. In order to determine the reliability of the two motion modes of the finger, this section analyzes the elastic mechanism and establishes the kinematic model of the dexterous finger in series with the elastic element. 

\begin{figure}
    \centering
    \includegraphics[width=1\linewidth]{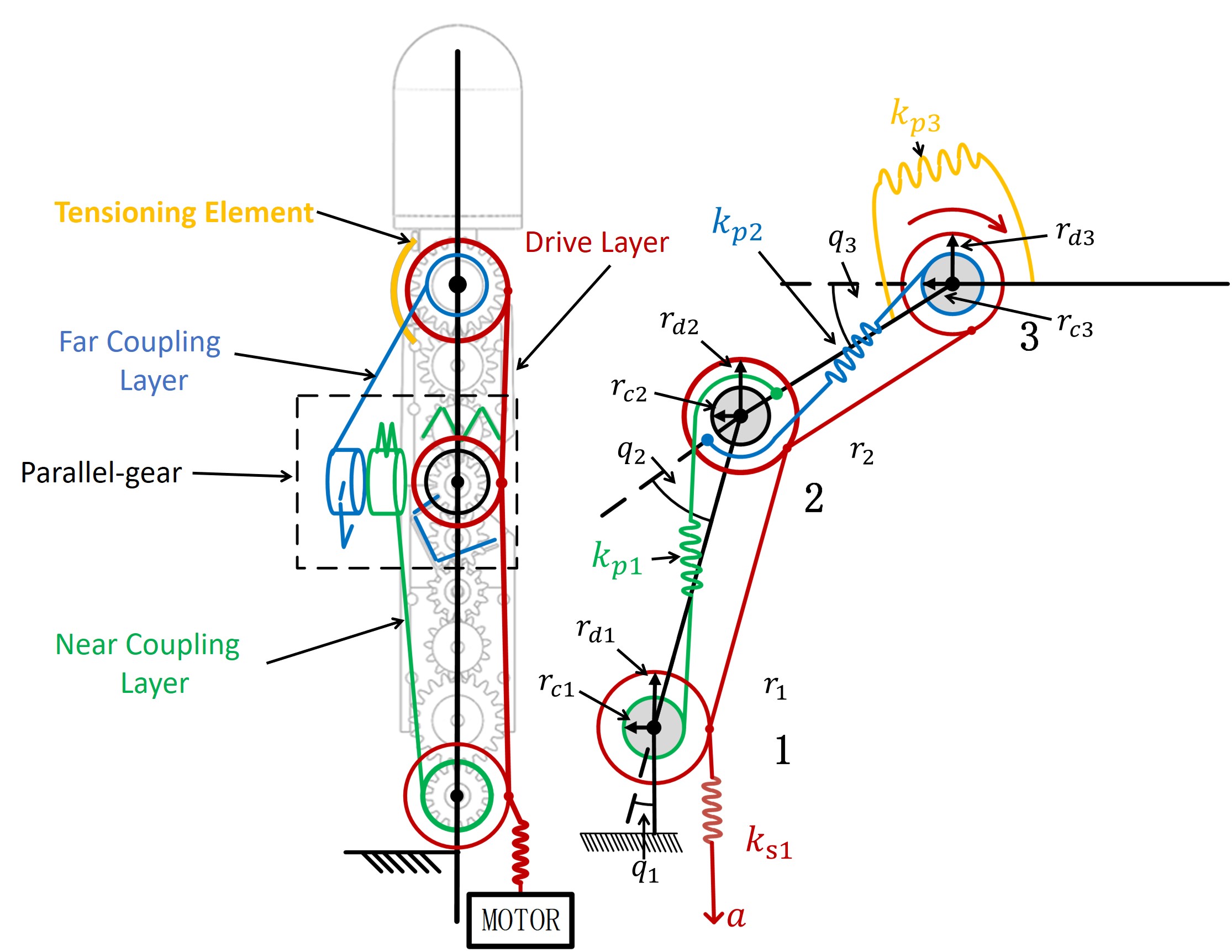}
    \caption{The adaptive UCM kinematic model of the finger. Notably, for clarity in illustration, the idler gear set is abstractly depicted using line segments of distinct colors.}
    \label{fig:UCMS}
\end{figure}

The flexible coupling joint of the dexterous hand can be considered as a 3-DOF serial transmission underactuated mechanism. We introduce the UCM unified kinematics model \cite{chen2020analysis} for analysis, and the simplified model is shown in the Fig~\ref{fig:UCMS}.

The transmission variables of the system are expressed as
$$
\begin{cases}
T_s = a - r_{d1}q_1 - \frac{z_1}{z_2}r_{d2}q_2  - \frac {z_1}{z_3}r_{d3}q_3\\
T_{p1}=r_{c2}q_2 - r_{c1}q_1\\
T_{p2}=r_{c3}q_3 - r_{c2}q_2 \\
T_{p3}=r_{c3}q_3
\end{cases}
$$

Among them, $r_{d1},r_{d2},r_{d3}$ are the radius of the drive chain gears. $r_{c1},r_{c2},r_{c3}$ are the radius of the coupling chain gears. $q_{1},q_{2},q_{3}$ are joint motion angles, $z_1$, $z_2$, and $z_3$ denote the numbers of teeth of the gears in the drive chain, with their specific values listed in Table~\ref{tab:table1}. $T_s$ is the serial variable of the system, $T_p$ is the parallel variable of the system.

The derivative of the transmission variables is expressed as
$$\begin{bmatrix}
  \delta T_s\\\delta T_p
\end{bmatrix}=\begin{bmatrix}
  J_{s1} &J_{s2} \\
  J_p&0
\end{bmatrix}
\begin{bmatrix}
  \delta q\\\delta a
\end{bmatrix}$$

According to the transmission equation,the transmission Jacobian matrices are 
$$
\begin{cases}
	J_{s1}=\begin{bmatrix}
    -r_{d1}&-\frac{z_1}{z_2}r_{d2}&- \frac {z_1}{z_3}r_{d3}
\end{bmatrix} \\ \\
J_{s2}=1\\ \\
J_P=\begin{bmatrix}
- r_{c1} & r_{c2}&0\\
0 &- r_{c2}&r_{c3}\\
0&0&r_{c3}
\end{bmatrix}
\end{cases}
$$
Then
$$rank([J_{S}^T,J_P^T]) = 4$$
$$rank([J_{S}^T,J_P^T]) \geq 3$$

Since the number of constraints exceeds the number of joints, the transmission mechanism remains stable. It can be seen from the above formula that the transmission Jacobian matrix is determined by the structural parameters, independent of the joint angle and the drive coordinate, and the finger is a linear transmission mechanism. Therefore, the stiffness matrix of the system can be expressed as :
$$\begin{cases}
	\mathrm{K}_q = \mathrm{J}_{s1}^T \mathrm{K}_s \mathrm{J}_{s1} + \mathrm{J}_p^T \mathrm{K}_p \mathrm{J}_p \\
	\mathrm{K}_a = \mathrm{J}_{s2}^T \mathrm{K}_s \mathrm{J}_{s2} \\
	\mathrm{K}_{qa} = \mathrm{J}_{s1}^T \mathrm{K}_s \mathrm{J}_{s2}
\end{cases}
$$
 $\mathrm{K}_s=k_{s1},\mathrm{K}_p=diag(k_{p1},k_{p2},k_{p3})$ .
By substituting into the Jacobian matrix, the resulting stiffness matrix $K_q$ is obtained, which is positive definite. This indicates that the system can return to equilibrium once the external disturbance is removed.

According to the definition of UCM's motion. AMS (Active motion subspace) and  PMS (passive motion subspace) both are determined by  $-K_q^{-1}K_{qa}$. AMS is expressed as $\delta q_A$, where $\delta{a}$ is driving variable.

$$\delta q_A\;=\; \frac{K_q^{-1}\!\begin{bmatrix} r_{d1} & \tfrac{z_1}{z_2}r_{d2} & \tfrac{z_1}{z_3}r_{d3} \end{bmatrix}^\top} {\left\|K_q^{-1}\!\begin{bmatrix} r_{d1} & \tfrac{z_1}{z_2}r_{d2} & \tfrac{z_1}{z_3}r_{d3} \end{bmatrix}^\top\right\|}\delta{a}$$

PMS is expressed as $\delta q_P$ ,which is the constraint equation of passive motion.


\[
\delta q_P = \left\{ \begin{bmatrix} \delta q_1 \\ \delta q_2 \\ \delta q_3 \end{bmatrix} \in \mathbb{R}^3 \mid r_{c1}\frac{z_1}{z_3} \delta q_1 + r_{c2}\frac{z_2}{z_3} \delta q_2 + r_{c3} \delta q_3 = 0 \right\}
\]

The active force is expressed as
$$
\tau_A=
\begin{bmatrix}
    \frac{z1}{z3}r_{d1} & \frac{z2}{z3}r_{d2} & r_{d3}
\end{bmatrix}
$$

\begin{figure}[h]
    \centering    \includegraphics[width=1\linewidth]{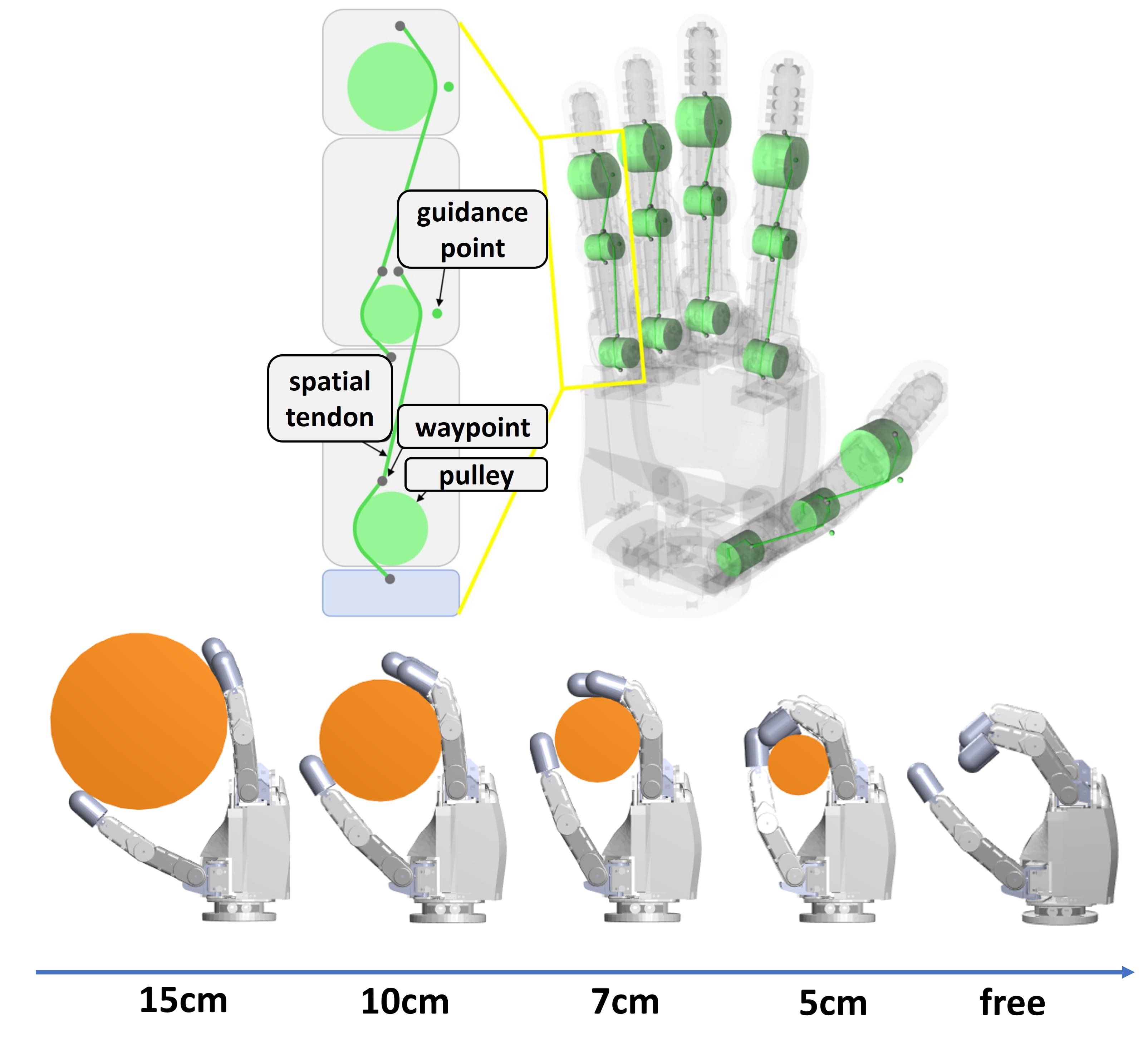}
    \caption{Equivalent simulation platform of the finger UCM, and adaptive enveloping experiments on spheres with different diameters.}
    \label{fig:自适应UCM验证}
\end{figure}

Thus, in the UCM system, through the optimization of spring stiffness parameters $K_s,K_p$, the coupled finger-to-finger motion and adaptive enveloping motion demonstrate feasible solutions within both the AMS and PMS. This configuration enables the system to maintain stable output of active extrusion forces onto the object, ensuring consistent grasping performance and adaptive manipulation capabilities. We reproduced this configuration in simulation and further validated the adaptive capabilities of the modular fingers through a set of grasping experiments. The results are presented in Fig.~\ref{fig:自适应UCM验证}.

\section{EXPERIMENTS}

In this section, modular fingers were developed based on the above analysis, and their performance was experimentally evaluated. Furthermore, a five-finger dexterous hand prototype was constructed using the modular fingers to assess overall performance.

\subsection{Modular Finger Kinematic Performance Evaluation}

\begin{figure}[h]
	\centering
	\includegraphics[width=1\linewidth]{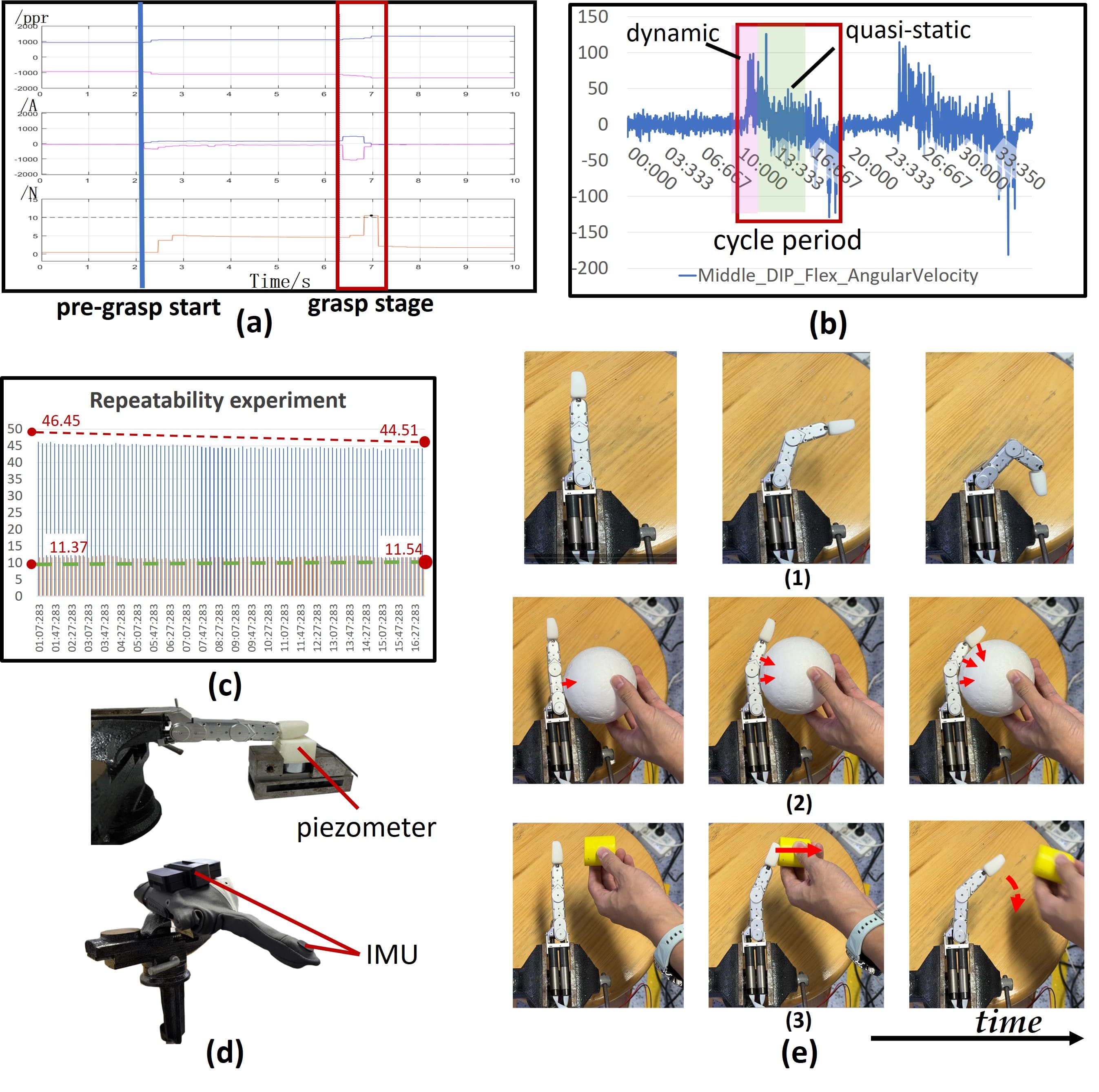}
	\caption{Modular finger mechanical performance experiment}
	\label{fig:modularfingertest}
\end{figure}

For kinematic performance assessment, we adopt the benchmarking framework proposed by Falco.\cite{falco2020benchmarking}. We developed a simplified testing platform comprising mechanical sensors and modular finger components.
In the kinematic performance evaluation experiment, a fixture was designed to secure the finger, integrating a uniaxial force sensor and an IMU sensor to capture finger orientation. This setup formed the experimental platform shown in Fig~\ref{fig:modularfingertest}(d). Experimental tests were conducted on grasp cycle time, finger repeatability, and fingertip force, with the results presented in Fig~\ref{fig:modularfingertest} (a), (b), and (c), respectively. The modular finger achieved a fingertip force of up to 10 N, and the repeatability performance is summarized in Table~\ref{tab:table3}.

\begin{table}[H]
	\centering
	\caption{FINGER REPEATABILIT\label{tab:table3}}
	 \resizebox{\linewidth}{!}{
	\begin{tabular}{|c|c|c|}
		\hline
		Avg Offset Distance(°) & Stdev(°) & 95\% Confidence Interval(°)\\\hline
		-0.8323 & 0.6266 & $\left( -0.9877, -0.6769 \right)$ \\\hline
	\end{tabular}
	}
\end{table}

The experimental validation of the adaptive UCM system is shown in Fig~\ref{fig:modularfingertest}(e): (1) During free motion, the elastic element acts as an inert gear, ensuring stable transmission and joint coupling. (2) In the ball-enveloping task, the proximal joint is constrained while the distal joint continues to flex until contact, after which all three joints apply compressive force for a stable grasp. Although the original coupling is disrupted, the system remains dynamically stable. (3) During fingertip contact, the object experiences vertical force until equilibrium is reached, and the mechanism returns seamlessly to its natural trajectory once the object is released.

\subsection{Dexterity Evaluation}
\begin{figure}
    \centering
    \includegraphics[width=1\linewidth]{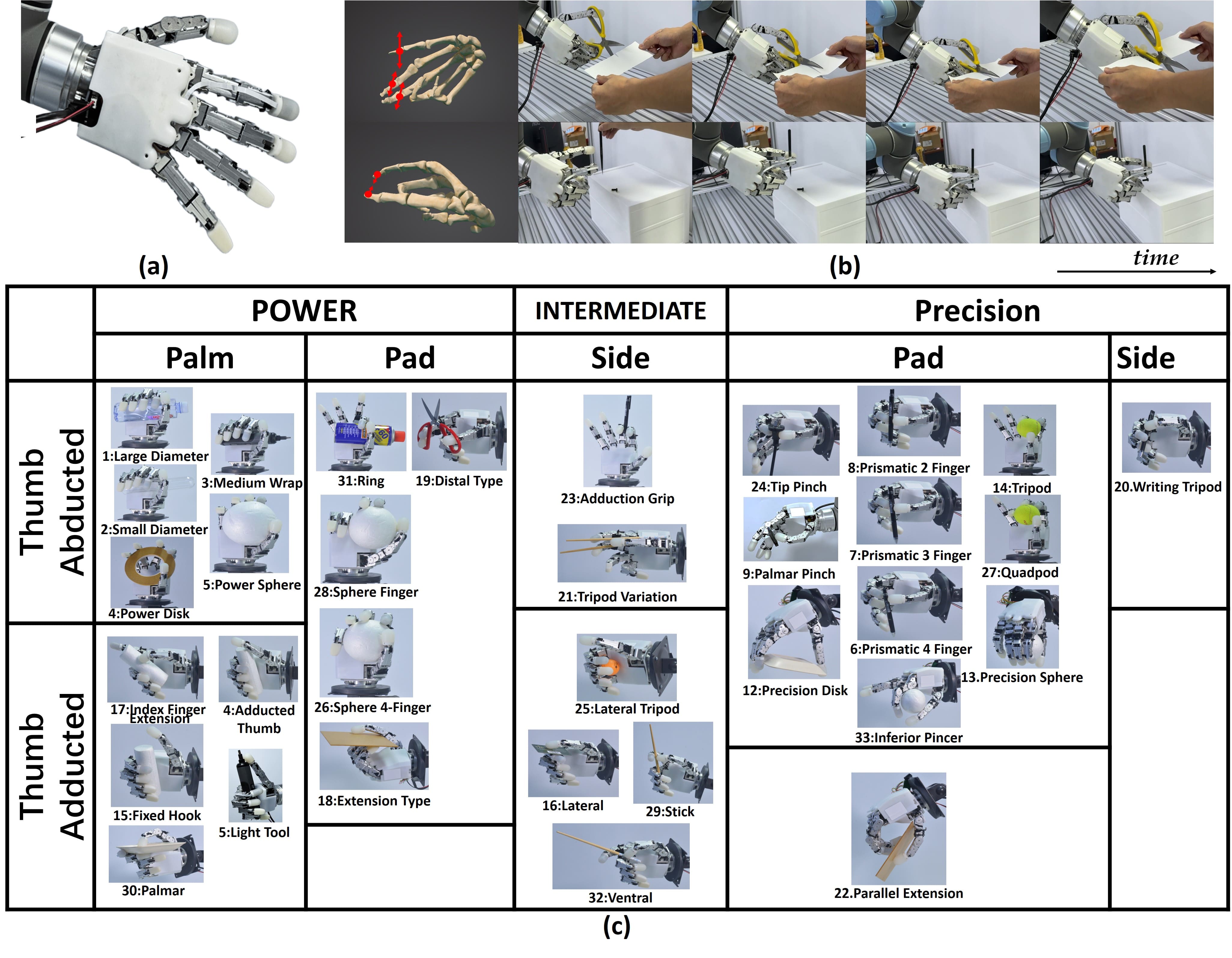}
    \caption{(a) Prototype of the modular five-finger dexterous hand; (b) Dexterity experiments in human-like tool–interaction scenarios; (c) Gesture mapping experiments based on the Feix grasp taxonomy.}
    \label{fig:灵巧性}
\end{figure}

We built a modular five-finger hand prototype as shown in Fig~\ref{fig:灵巧性}(a) to validate the design.Grasping ability is a critical metric for evaluating the Dexterity of dexterous hands. To assess the everyday grasping capability of the Modular Hand, we conducted experiments based on the Feix taxonomy\cite{feix2015grasp}. As shown in Fig~\ref{fig:灵巧性}(c), the proposed dexterous hand successfully performed all 33 gestures. There is currently no universally accepted quantitative standard for in-hand manipulation dexterity. Following the cross-disciplinary view that tool use is a defining marker of advanced manual skill \cite{marzke1997precision}\cite{qin2023robot}, we conducted tool-use experiments to evaluate the anthropomorphic manipulation capability of the proposed dexterous hand. We conducted sequential manipulation experiments with scissors and tweezers, as illustrated in Fig~\ref{fig:灵巧性}(c).

\section{CONCLUSIONS}

Grounded in human hand anatomy and synergy-based posture synthesis, this study presents an anthropomorphic five-finger topology and designs an adaptive, modular dexterous hand. The performance of the modular finger was analyzed and a physical prototype was constructed for validation. The results of the experiment demonstrate strong dexterity in both grasping and in-hand manipulation.

Future investigations will focus on preserving biomimetic manipulation capabilities under fewer actuation schemes. In parallel, efforts will be directed toward mechanism parameter optimization, tactile sensing integration, and the formulation of task-specific control algorithms. These efforts expected to advance grasp stability, manipulation precision, and the overall functional versatility of dexterous robotic hands. The modular design provides a flexible foundation for subsequent research and development.






\newpage
\bibliographystyle{IEEEtran}  
\bibliography{ref}            
\end{document}